\documentclass[10pt,twocolumn,letterpaper]{article}

\usepackage[pagenumbers]{cvpr} 

\usepackage{graphicx}
\usepackage{amsmath}
\usepackage{amssymb}
\usepackage{booktabs}
\usepackage{ragged2e} 
\usepackage{makecell, multirow, tabularx}
\usepackage[table]{xcolor}

\usepackage[pagebackref,breaklinks,colorlinks]{hyperref}

\usepackage[capitalize]{cleveref}
\crefname{section}{Sec.}{Secs.}
\Crefname{section}{Section}{Sections}
\Crefname{table}{Table}{Tables}
\crefname{table}{Tab.}{Tabs.}

\def\cvprPaperID{3794} 
\def\confName{CVPR}
\def\confYear{2023}

\begin{document}

\title{Focus On Details:\\Online Multi-object Tracking with Diverse Fine-grained Representation}
 
\author{
Hao Ren \textsuperscript{1}, Shoudong Han \textsuperscript{1, $\ast$} , Huilin Ding, Ziwen Zhang, Hongwei Wang, Faquan Wang  \\
Huazhong Univerisity of Science and Technology \\
\tt\small \{haoren2000, shoudonghan\}@hust.edu.cn}

\maketitle
\renewcommand{\thefootnote}{}
\footnotetext{\textsuperscript{1} Equal contribution} 
\footnotetext{\textsuperscript{$\ast$} Corresponding author} 

\begin{abstract}
Discriminative representation is essential to keep a unique identifier for each target in Multiple object tracking (MOT). Some recent MOT methods extract features of the bounding box region or the center point as identity embeddings. However, when targets are occluded, these coarse-grained global representations become unreliable. To this end, we propose exploring diverse fine-grained representation, which describes appearance comprehensively from global and local perspectives. This fine-grained representation requires high feature resolution and precise semantic information. To effectively alleviate the semantic misalignment caused by indiscriminate contextual information aggregation, Flow Alignment FPN (FAFPN) is proposed for multi-scale feature alignment aggregation. It generates semantic flow among feature maps from different resolutions to transform their pixel positions. Furthermore, we present a Multi-head Part Mask Generator (MPMG) to extract fine-grained representation based on the aligned feature maps. Multiple parallel branches of MPMG allow it to focus on different parts of targets to generate local masks without label supervision. The diverse details in target masks facilitate fine-grained representation. Eventually, benefiting from a Shuffle-Group Sampling (SGS) training strategy with positive and negative samples balanced, we achieve state-of-the-art performance on MOT17 and MOT20 test sets. Even on DanceTrack, where the appearance of targets is extremely similar, our method significantly outperforms ByteTrack by 5.0\% on HOTA and 5.6\% on IDF1. Extensive experiments have proved that diverse fine-grained representation makes Re-ID great again in MOT.
\vspace{-0.6cm}
\end{abstract}

\begin{figure}
	\setlength{\abovecaptionskip}{0.0cm}
	\begin{minipage}{0.4\linewidth}
		\centering
		\begin{subfigure}{1\linewidth}
			\centering
			\includegraphics[width=12mm]{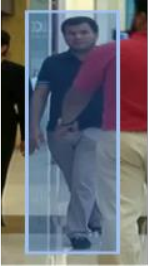}
			\caption{Bbox-based}
			\label{fig:Bbox-based}
		\end{subfigure}
		\begin{subfigure}{1\linewidth}
			\centering
			\includegraphics[width=12mm]{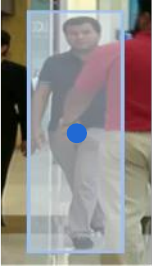}
			\caption{Center-based}
			\label{fig:Center-based}
		\end{subfigure}
	\end{minipage}
	\medskip
	\begin{minipage}{0.6\linewidth}
		\centering
		\begin{subfigure}{1\linewidth}
			\centering
			\includegraphics[width=52mm]{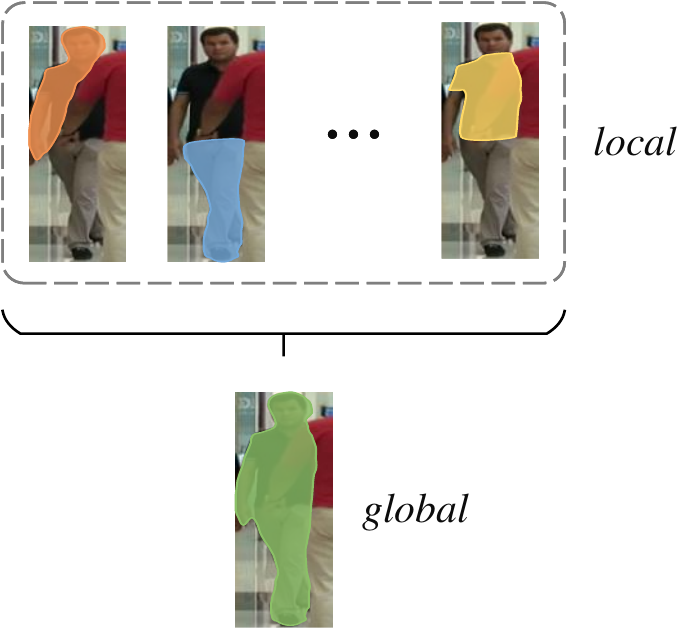}
			\caption{Ours}
			\label{fig:Ours}
		\end{subfigure}
	\end{minipage} 
	\caption{{\bf Comparison with different methods of appearance representation}: (a) Bbox-based, (b) Center-based, (c) global-local fine-grained representation (ours).}
	\label{fig:compare}
\end{figure}

\section{Introduction}
\label{sec:intro}
As a fundamental task in computer vision, multi-object tracking (MOT) is crucial for automatic driving, video surveillance, etc. MOT aims to localize targets and maintain their unique identities. Recent MOT methods\cite{bochinski2017high,sort,yu2016poi,MTrack,bytetrack} mainly follow the paradigm of tracking-by-detection, and divide tracking into two independent steps: detection and association. The detector detects targets in each frame first, and then appearance representation and position information are employed as the association basis to link targets with the corresponding trajectories. As the inherent attributes of the target, appearance and position complement each other in the association. 
\par
However, due to targets or camera motion, intra-class and inter-class occlusion are inevitable, which puts forward stricter requirements for appearance representation. As shown in \cref{fig:Bbox-based} and \cref{fig:Center-based},  these methods extract the features of the bounding box region or the center point as appearance embeddings. However, these coarse-grained global embeddings are extremely sensitive to noise, so that become unreliable once the signal-to-noise ratio is reduced. With the headway of the detector\cite{ren2015faster,fcos,zhou2019objects,carion2020end,yolox}, appearance representation gradually cannot keep the same performance as detection. Some researchers\cite{bytetrack} find that simply using position cues is enough to obtain satisfactory results, while appearance cues are unfavorable for further improvement.
\par
To break this situation, we re-examine the recent appearance-based methods. The bbox-based methods\cite{voigtlaender2019mots} in \cref{fig:Bbox-based} adopt global average pooling, which converts features of the bounding box region into appearance embeddings. These methods equally treat target and interference features (background and other objects), which is unreasonable. As shown in \cref{fig:Center-based}, researchers\cite{centertrack,fairmot,relationtrack} notice this issue and utilize features at the target centers as their appearance embeddings, eliminating interference from the background as much as possible. Despite this, when the target is occluded, the feature at its center is still inevitably interfered with noise information from other objects. The fuzziness of global representation has become an impediment of these methods. On the contrary, our method focuses on different local details of targets, which is illustrated in \cref{fig:Ours}. Fine-grained global and local representations complement each other and jointly describe appearance. When the target is occluded, our method can still identify it according to visible parts, similar to human judgment. 
\par
As the basis of fine-grained representation, target feature maps require high-resolution and unambiguous semantic information. Shallow or deep outputs cannot meet these two requirements simultaneously. Therefore, it is feasible to enrich the semantic information of shallow features or improve the resolution of deep features. To reduce the burden, researchers usually adopt FPN\cite{FPN} to aggregate multi-scale shallow feature maps indiscriminately, which causes semantic misalignment among features with different resolutions. Specifically, there is a spatial misalignment between the late feature maps after up-sampling and the early feature maps. 
\par
To solve this issue, we construct a \textit{Flow Alignment FPN} (FAFPN) to learn the semantic flow among feature maps with different scales and effectively broadcast high-level features to high-resolution features. FAFPN aligns feature maps by semantic flow and aggregates context information to enrich semantic information while maintaining high resolution. Further, \textit{Multi-head Part Mask Generator} (MPMG) is proposed to generate part masks for detailed representations without label supervision. Inspired by multi-head self-attention in Transformer\cite{Transformer}, MPMG implements a multi-branch parallel structure, which can efficiently and comprehensively focus on different parts of the target. Combining FAFPN and MPMG, we can obtain a diverse fine-grained representation, including diverse local embeddings and background-filtered global embeddings.
\par
In the training phase, some current MOT methods\cite{fairmot,liang2022rethinking} train Re-ID (re-identification) by following the video sequence or shuffling all video frames. The former does not disperse training data, while the latter is positive and negative samples imbalanced. To train Re-ID more reasonably, we propose a training strategy named \textit{Shuffle-Group Sampling} (SGS). In this strategy, we group video frames into short segments in their order and then shuffle these segments. SGS disperses data and balances positive and negative samples. Our model incorporates all the above proposed techniques, named \textit{fine-grained representation tracker} (FineTrack).
\par 
The main contributions of our work can be summarized as follows:
\begin{itemize}
\setlength{\itemsep}{0 pt}
\setlength{\parsep}{0 pt}
\setlength{\parskip}{0 pt}
	\item We propose a \textit{Flow Alignment FPN} (FAFPN). It learns the semantic flow among feature maps with different resolutions to correct spatial dislocation. Feature maps after FAFPN include high resolution and precise semantic information, which is the basis of fine-grained representation.
	\item We construct a \textit{Multi-head Part Mask Generator} (MPMG) to focus on details of targets. MPMG employs self-attention to filtrate background noise and extract global-local embeddings to represent targets comprehensively.
	\item \textit{Shuffle-Group Sampling} (SGS) is proposed to disperse training data and balance positive and negative samples. It reduces oscillation in model convergence to achieve better performance.
\end{itemize}
\begin{figure*}
	\centering
	\includegraphics[width=175mm]{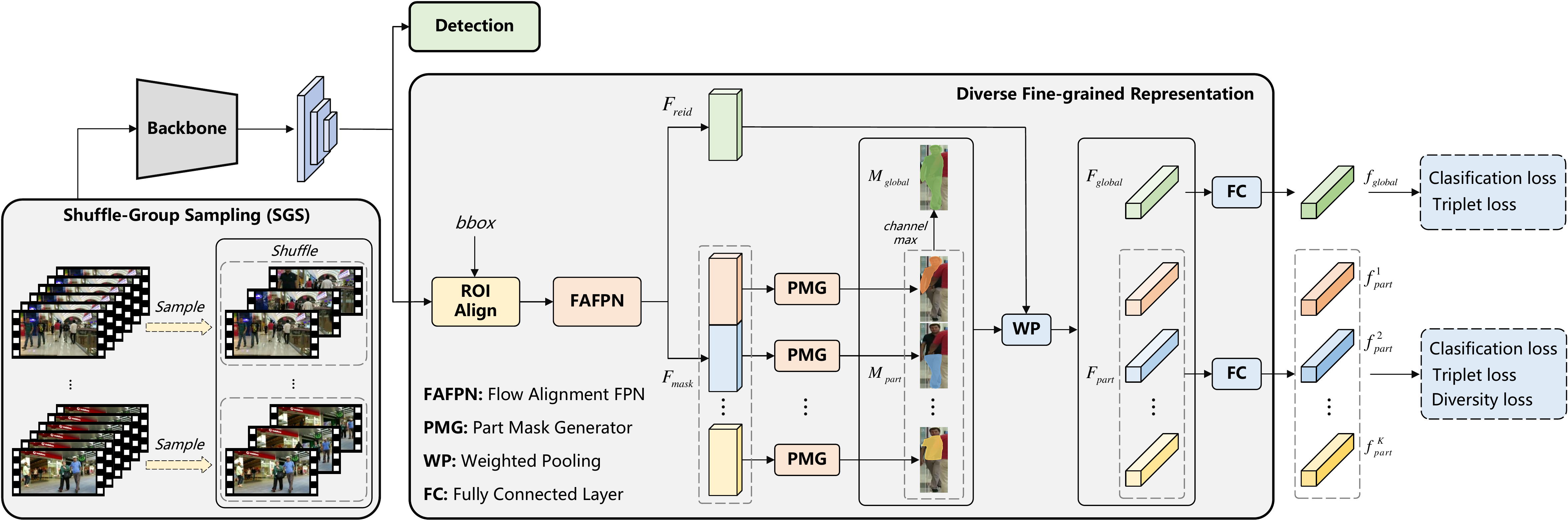}
	\caption{{\bf Overall pipeline of FineTrack.} FineTrack comprises 5 steps: (1) SGS divides videos into groups and shuffles them. (2) Extract feature maps of the input frame for detection. Meanwhile, multi-scale shallow feature maps are obtained for ROIAlign to output target feature maps with different resolutions. (3) FAFPN aligns target feature maps and aggregates their context information for ${F_{mask}}$ and ${F_{reid}}$. (4) Divide ${F_{mask}}$ into ${K}$ blocks along their channel and employ multi-head PMG to generate part masks ${M_{part}}$. Concatenate  ${M_{part}}$ and adopt ${Max}$ of channel to form global masks ${M_{global}}$. (5) ${F_{global}}$ and ${F_{part}}$ can be obtained by a weighted pooling. Then two fully connected layers convert them into global and part embedding: ${f_{global}}$ and ${f_{part}}$, respectively.}
	\label{fig:overview}
\end{figure*}
\section{Related Work}
\noindent 
{\bf Tracking-by-detection.} With the development of detection, many works\cite{bergmann2019tracking,tubetk,chained,wang2020towards} adopt the tracking-by-detection paradigm, and divide MOT into detection and association.The trackers following this paradigm first detect objects in video frames and then associate them dependent on their identity information. Some works\cite{bergmann2019tracking,feichtenhofer2017detect,siammot,DAN} model object movement for identity association. SORT\cite{sort} relies on the Kalman Filter\cite{kalman} to predict future positions of the targets, calculates their overlap with detection, and utilizes the Hungarian algorithm\cite{kuhn1955hungarian} for identity association. When the target moves irregularly or is occluded, only using motion information is not enough to achieve better performance. Based on SORT\cite{sort}, DeepSORT\cite{deepsort} introduces the appearance information of targets to enhance identity embedding. 
\par 
Appearance and motion, as inherent properties of targets, are not independent but complementary while tracking. However, appearance does not always provide reliable clues for identification, particularly when targets are occluded. Our method focuses on diverse details of targets and constructs global-local fine-grained representation, which achieves better performance in occlusion scenes.
\\
{\bf Re-ID in MOT.} Some works\cite{zhou2018online,fang2018recurrent,yu2016poi} crop the image regions of detections and extract features with an extra Re-ID model. These works treat detection and Re-ID as independent tasks, imposing additional parameter burdens. To solve this problem, JDE\cite{wang2020towards} and FairMOT\cite{fairmot} implement a branch the same as detection for appearance representation. The structure of the multi-task branches achieves impressive performance but also raises an additional issue: competition caused by different goals for detection and Re-ID. To alleviate this conflict, CSTrack\cite{liang2022rethinking} constructs a CCN module to extract general and specific features more suitable for detection and Re-ID. However, these appearance-based methods extract the features of the boundary box region or the center point as appearance representation, which are inevitably interfered with irrelevant information. They focus on generating more discriminative global embedding instead of exploring more detailed appearance cues.
\par
Although well-designed Re-ID modules can improve performance, global features are still prone to ambiguity. With a high-performance detector, the associator can link most targets only with motion cues. The global representation is also impotent for the remaining targets incorrectly tracked due to occlusion or other factors. Using global appearance cues at this point can even compromise tracking accuracy. Instead, our method focus on the details of targets. Following this idea, we construct the FAFPN to align and aggregate high-resolution shallow feature maps and feed them into the MPMG to obtain fine part masks. Further,  part masks of targets are employed to extract the fine-grained global-local appearance embedding, which can more comprehensively represent the identity.
\section{Methodology}
\subsection{Overview}
\label{sec:3.1}
In this work, we adopt the high-performance detector YOLOX\cite{yolox} to detect targets. As shown in  \cref{fig:overview}, during training, SGS groups video frames by sampling them sequentially, ensuring targets with positive and negative samples. After shuffling, these grouped data are fed into the Backbone to obtain feature maps with different resolutions. Then, we adopt ROIAlign to crop and scale feature maps of the bbox regions to get the multi-scale feature maps of targets. Immediately, FAFPN aligns these target feature maps from different resolutions and aggregates them into fine-grained feature maps: ${F_{mask}}$ and ${F_{reid}}$ for mask generation and appearance embedding, respectively. After slicing  ${F_{mask}}$ along the channel dimension, the features are fed into separate \textit{Part-Mask Generator} (PMG) to generate target masks. Further, ${F_{global}}$ and ${F_{part}}$ can be obtained by a weighted pooling. These embeddings are fed into two fully connected layers that do not share parameters to output ${f_{global}}$ and ${f_{part}}$. For ${f_{global}}$, we calculate Classification loss and Triplet loss. For ${f_{part}}$, a diversity loss is added to expand the discrepancy among different part features.
\subsection{Flow Aligned FPN}
\label{sec:3.2}
As a classical method for feature aggregation, FPN\cite{FPN} has been extensively applied in different computer vision tasks. However, step-by-step downsampling and indiscriminate context aggregation cause semantic misalignment among feature maps of different scales. Fuzzy semantic information have a significant impact on visual tasks which require detailed descriptions. To obtain fine-grained features, we employ the \textit{Flow Alignment Module} (FAM)\cite{semanticflow,alignseg} to generate semantic flow among feature maps of different resolutions. The semantic flow can guide alignment and eliminate spatial dislocation among feature maps from different scales. Furthermore, we utilize the FAM to optimize the aggregation process of FPN and then construct a \textit{Flow Alignment FPN} (FAFPN). 
\par
\begin{figure}
	\centering
	\includegraphics[width=70mm]{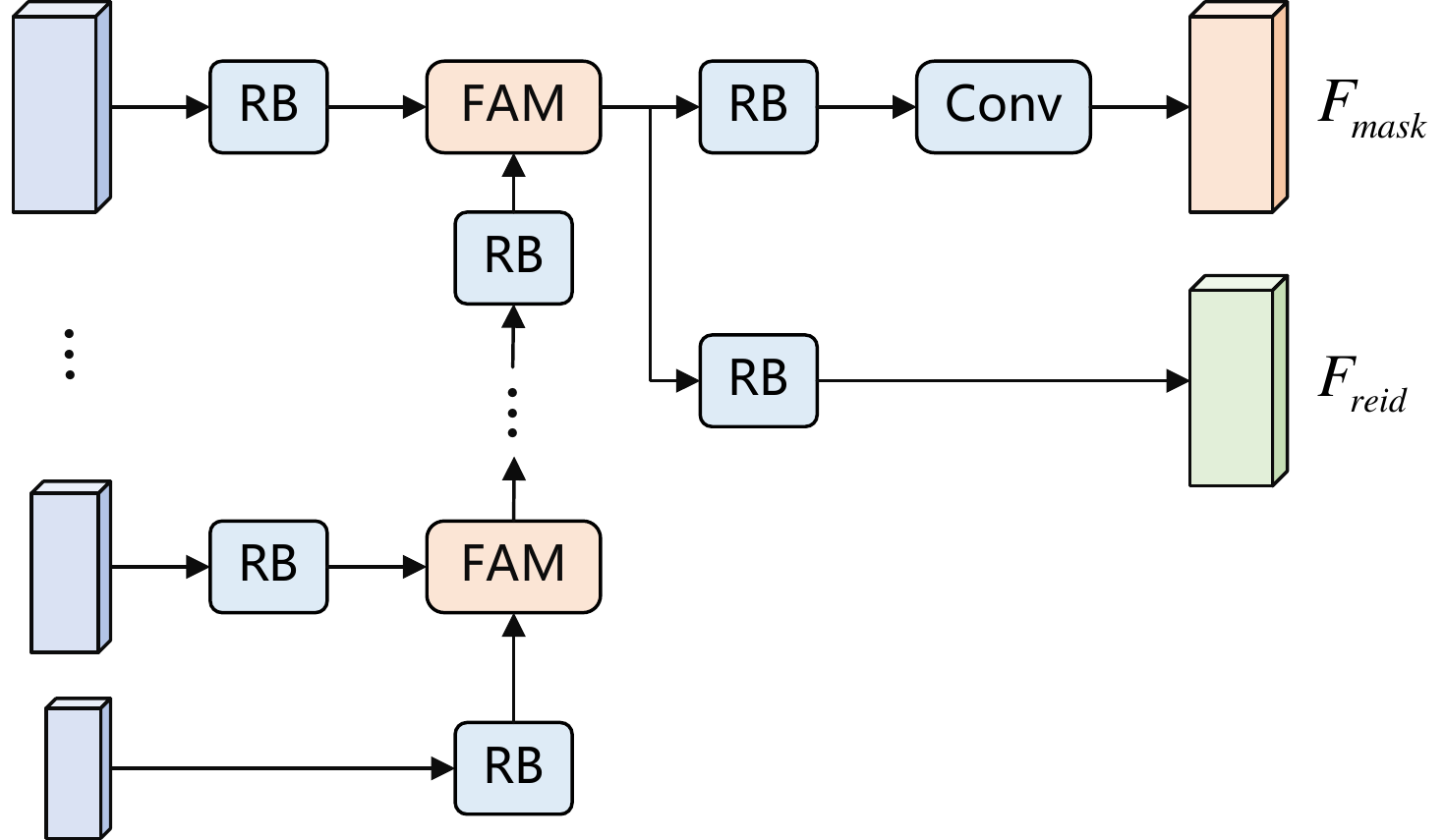}
	\caption{{\bf Structure of FAFPN.} FAFPN consists of FAM and RB for aggregating feature maps from $ S $ different scales}
	\label{fig:FAFPN}
\end{figure}

\begin{figure}
	\setlength{\abovecaptionskip}{0.2cm}
	\begin{subfigure}{0.3\linewidth}
		\centering
		\includegraphics[width=18mm]{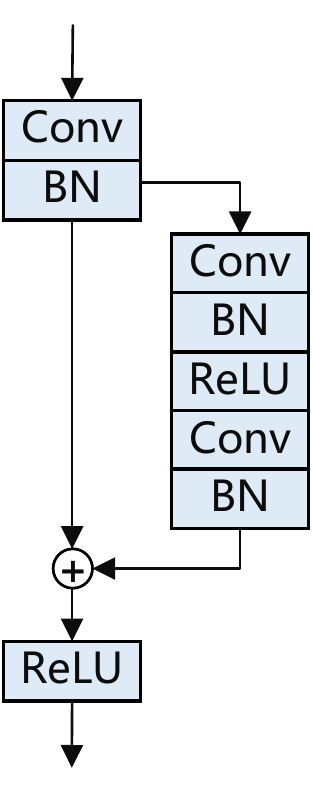}
		\caption{RB}
		\label{fig:RB}
	\end{subfigure}
	\hfill
	\begin{subfigure}{0.7\linewidth}
		\centering
		\includegraphics[width=40mm]{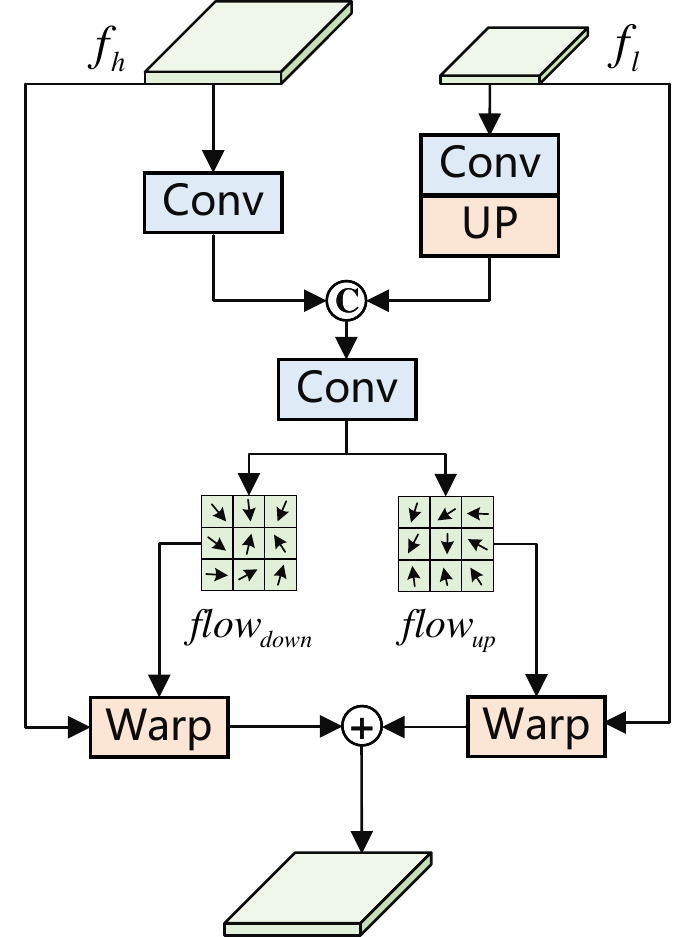}
		\caption{FAM}
		\label{fig:FAM}
	\end{subfigure}
	\caption{{\bf Implementation of RB and FAM.} RB is a form of residual structure. FAM learns semantic flow to align feature maps from different scales.}
	\label{fig:RB&FAM}
\end{figure}

 
The overall structure of FAFPN is shown in \cref{fig:FAFPN}. \textit{ResBlock} (RB) replaces ${1*1}$ Convolution in FPN to fuse shallow features more adequately. The $ Sum $ operation is replaced with the FAM, making semantic alignment possible. The multi-scale feature maps are modified to a unified channel dimension by RB and then fed into the FAM for alignment and aggregation. Finally, we use two RB to generate ${F_{mask}}$ and ${F_{reid}}$ respectively for meeting the different requirements of the mask and representation. \cref{fig:RB&FAM} shows the specific structure of RB and FAM. The RB is a simple residual block. It combines ${1*1}$ Convolution and BN layer to transform the input channel dimension and then applies the residual structure to get its output. The FAM takes two feature maps from different scales as input. ${f_{l}}$ has the same size as ${f_{h}}$ after up-sampling. Further, they are concatenated to generate semantic flow: ${flow_{down}}$ and ${flow_{up}}$, which represent the pixel offset of feature maps ${f_{h}}$ and ${f_{l}}$, respectively. $ Warp $ operation aligns these feature maps based on semantic flow. In the end, the FAM employs an $ Sum $ operation to aggregate the context information of aligned feature maps.
\par
After semantic alignment, multi-scale shallow feature maps are aggregated into high-resolution feature maps with precise semantic information, which are the basis for generating diverse fine-grained representations.
\subsection{Multi-head Part Mask Generator}
\label{sec:3.3}
Some recent works rely on well-designed modules to optimize global embedding. These methods indiscriminately generalize the various details of the target into a coarse-grained global embedding, which causes semantic ambiguity. When the target is occluded, the features of the bounding box region or the center point are inevitably disturbed by noise. 
\par
\begin{figure}
	\centering
	\includegraphics[width=85mm]{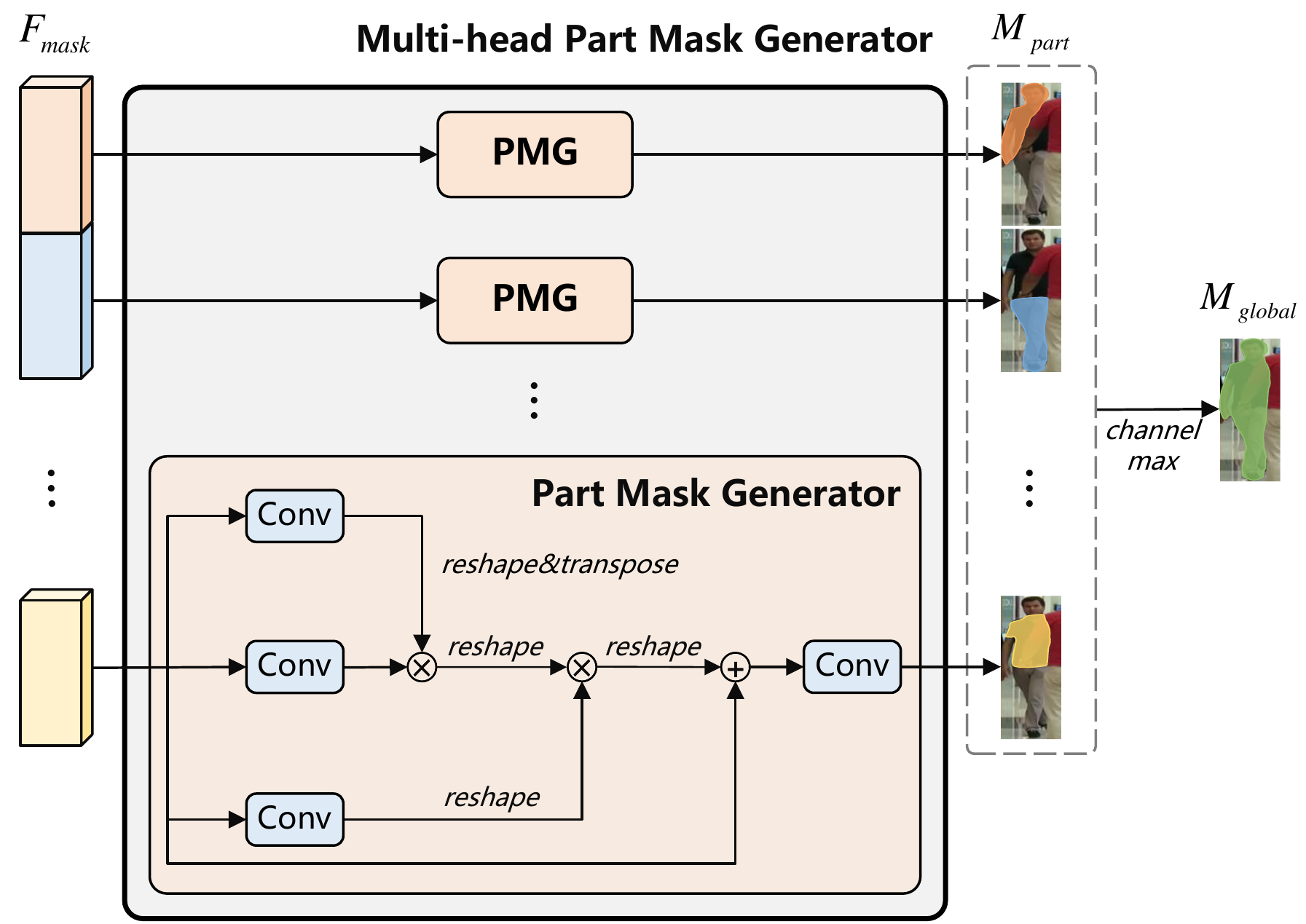}
	\caption{{\bf Illustration of MPMG.} Multiple parallel PMG branches focus on different parts of the target and generate part masks.}
	\label{fig:MPMG}
\end{figure}

To alleviate this problem, we propose exploring fine-grained appearance representation to focus on target diversity details. The direct approach is to adopt an attention module to enhance feature maps. There are many excellent attention modules in Convolutional Neural Networks, such as Non-Local\cite{Non-local}, CBAM\cite{cbam}, etc. Simply employing attention modules can not entirely focus on the details of targets. Inspired by the multi-head self-attention module in Transformer\cite{Transformer}, we establish multiple parallel attention branches to focus on different parts of targets. At the same time, we employ the Non-Local block as the branch attention module, followed by a Convolution to generate a mask of the target part. As shown in \cref{fig:MPMG}, we divide the input feature maps into blocks along their channel dimension, and each block can be considered as a different mapping of the target feature. These feature maps are fed into \textit{Multi-head Part Mask Generator} (MPMG) to generate different part masks of the target without Ground Truth. 
\par
For any branch in MPMG namely PMG, three ${1*1}$ Convolutions are respectively used to obtain query, key, and value. Query and key generate the attention matrix, which is multiplied by value and then added with input. After that, a layer combining ${1*1}$ Convolution and Sigmoid is used to generate a part mask with values between 0 and 1. The same operation is performed for the remaining feature blocks, resulting in ${K}$ different part masks. After concatenation, the channel of part masks is aggregated by ${Max}$ to form a global mask containing different local information. It is worth noting that parameters are not shared among parallel branches, which ensures sufficient diversity of attention.
\subsection{Train and Inference}
\label{sec:3.4}
\noindent 
{\bf Shuffle-Group Sampling.} For better detection performance, researchers shuffle consecutive video frames, which is effective for training detection but inappropriate for training Re-ID. This training strategy makes a batch contain almost no targets with the same identity. Such imbalanced positive and negative samples cause the model only to discover the difference among the different targets but fail to capture the commonalities of the same targets. This is particularly detrimental to Re-ID, which aims to distinguish the identity of targets.
\par
To solve the above problem, we construct the \textit{Shuffle-Group Sampling} (SGS) training strategy. Different from random sampling, SGS adopts sequential sampling to group video frames. In this way, targets in the same batch hold positive samples with the same identity, thus alleviating the problem of imbalanced positive and negative samples caused by random sampling. In training, we disrupt the order of grouped data to reduce convergence fluctuations. This strategy allows contiguous batches to originate from different videos and form a sequence of track segments with significant appearance variations. Such a discrete distribution of training data optimizes the convergence direction of the model.
\\
{\bf Training Loss.} 
SGS enables a batch of targets to contain both positive and negative samples, so the model can be optimized using Triplet loss, which is extensively applied in Re-ID methods. The part and global features are denoted as ${f_{part}}=\left\{ {f_{p}^{n,k},n \in \left[ {1,2,...,N} \right],k \in \left[ {1,2,...,K} \right]} \right\}$ and ${f_{global}} = \left\{ {f_{g}^n,n \in \left[ {1,2,...,N} \right]} \right\}$. Here $N$ represents the number of targets in the image and each target is composed of $K$ parts. Thus, $f_{p}^{n,k}$ means the ${k_{th}}$ part feature of the ${n_{th}}$ target, and $f_{g}^n$ represents the global feature of the ${n_{th}}$ target.
\par
Then we can calculate the Triplet loss with a soft margin for part features based on \cref{eq:2}:
\begin{equation}
	L_{tri}^{p}(k) = \frac{1}{K}\sum\limits_{k = 1}^K {Triplet(f_{p}^{k})},
	\label{eq:2}
\end{equation}
where $Triplet(\cdot)$ is the same as \cite{transreid} and $f_{p}^{k}$ represents the ${k_{th}}$ part features of $n$ targets in the image.
\par
Similarly, the Triplet loss of the global features is based on \cref{eq:3} :
\begin{equation}
	L_{tri}^{g}(k) = {Triplet(f_{global})},
	\label{eq:3}
\end{equation}
\par
After obtaining $K$ part features, $K$ combinations of Linear layer and Softmax are respectively applied to get the classification result vectors $P = \left\{ {p_n^k,k \in \left[ {1,2,...,K} \right],n \in \left[ {1,2,...,N} \right]} \right\}$. The meanings of $K$ and $N$ are the same as in ${f_{part}}$. $p_n^k$ represents the classification vector of the ${k_{th}}$ part feature  of the ${n_{th}}$ target. The dimension of $p_n^k$ is $M$ , which is also the number of all targets in the datasets. For target identity labels, we denote them as $Y = \left\{ {{y_{n,m}},n \in \left[ {1,2,...,N} \right],m \in \left[ {1,2,...,M} \right]} \right\}$, ${y_{n,m}}$ means whether the ${n_{th}}$ target has the same identity as the ${m_{th}}$ target in ID labels, with a value of 0 or 1. According to the above definition, we can calculate the classification loss of the ${k_{th}}$ part feature of the ${n_{th}}$ target with \cref{eq:4}:
\begin{equation}
	L_{n,k}^p(m) = {y_{n,m}}\log (p_n^k(m)),m \in [1,2,...,M]
	\label{eq:4}
\end{equation}
\par
Further, we calculate the classification loss of part features based on with \cref{eq:5}:
\begin{equation}
	L_{cls}^p =  - \frac{1}{K \cdot N}\sum\limits_{k = 1}^K {\sum\limits_{n = 1}^N {\sum\limits_{m = 1}^M {L_{n,k}^p(m)} } }
	\label{eq:5}
\end{equation}
\par
Similarly, the classification loss of global features is:
\begin{equation}
	L_{cls}^g =  - \frac{1}{N} \sum\limits_{n = 1}^N {\sum\limits_{m = 1}^M {{y_{n,m}}\log ({g_n}(m))}},
	\label{eq:6}
\end{equation}
where ${g_n}$ is the classification vector of the ${n_{th}}$ target's global feature.
\par
Due to the peculiarities of multi-branch structures, using only classification loss and Triplet loss does not ensure that the model focuses on different parts of the target. To avoid multiple branches gazing at similar details, we employ the diversity loss $L_{div}$ in \cref{eq:7} to distance different part features of the same target:
\begin{equation}
	{L_{div}} = \frac{1}{{N \cdot K(K - 1)}}\sum\limits_{n = 1}^N {\sum\limits_{{k_i} \ne {k_j}}^K {\frac{{\left\langle {f_p^{n,{k_i}},f_p^{n,{k_j}}} \right\rangle }}{{{{\left\| {f_p^{n,{k_i}}} \right\|}_2} \cdot {{\left\| {f_p^{n,{k_j}}} \right\|}_2}}}} }
	\label{eq:7}
\end{equation}
\par
The purpose of diversity loss is intuitive, which is to keep the cosine similarity between different part features of the same target as low as possible.
\par
We combine the above losses into a final training loss:
\begin{equation}
	L = \alpha  \cdot (L_{cls}^p + L_{tri}^p) + \beta  \cdot (L_{cls}^g + L_{tri}^g) + \gamma  \cdot {L_{div}},
	\label{eq:8}
\end{equation}
where $\alpha,\beta$ and $\gamma$ are used to adjust the proportion of different losses and we set $\alpha  = 3,\beta  = 0.3,\gamma  = 2$.
\\
{\bf Inference.} 
Based on ByteTrack\cite{bytetrack}, we add a method similar to DeepSort\cite{deepsort} that calculates Re-ID features into feature distance matrix. It is worth mentioning that we concatenate part features of targets with global features as Re-ID features.
\par
We adopt the exponential moving average (EMA) mechanism to update the features $\tilde f_i^t$ of matched tracklets for the $i_{th}$ tracklet at frame $t$ based on \cref{eq:ema}, as in \cite{fairmot}.
\begin{equation}
	\tilde f_i^t = \lambda \tilde f_i^{t - 1} + (1 - \lambda )f_i^t,
	\label{eq:ema}
\end{equation}
where $f_i^t$ is the feature of the current matched detection, and $ \lambda = 0.9 $ is a momentum term.
\par
In the tracking algorithm, the feature distance matrix ${d_{feat}}$ is:
\begin{equation}
	{d_{feat}} = 1 - Similarity(\tilde f^{t - 1}, f^t),
\end{equation}
where $Similarity(\cdot)$ outputs the cosine similarity matrix between tracklets features $\tilde f^{t - 1}$ and targets features $f^t$.
\par
Meanwhile, we can calculate the IoU distance matrix ${d_{IoU}}$ base on \cref{eq:9}.
\begin{equation}
	{d_{IoU}} = 1 - IoU({b_{det}},{b_{pre}}),
	\label{eq:9}
\end{equation}
where $IoU(\cdot)$ outputs the IoU matrix between detection bboxes $ b_{det} $ and prediction bboxes $ b_{pre} $.
\par
To exclude the interference of distant targets, we only consider feature distances between pairs of targets with the IoU distance less than 1, which means there is bounding box overlap. The optimized feature distance matrix ${\tilde d_{feat}}$ is:
\begin{equation}
	{\tilde d_{feat}} = 1 - (1 - {d_{feat}}) \cdot ({d_{IoU}} < 1)
\end{equation}
\par
After squaring the product of the optimized feature distance matrix and IoU matrix, the final distance matrix is obtained for the association with \cref{eq:13}:
\begin{equation}
	d = \sqrt {{\tilde d_{feat}} \cdot {d_{IoU}}}
	\label{eq:13}
\end{equation}
\par 
Finally, we set the association threshold to 0.5.
\begin{table} \small
	\centering
	\tabcolsep=0.4mm
	\begin{tabular}{@{}lcccccc@{}}
		\toprule
		Method & HOTA$\uparrow$ & IDF1$\uparrow$ & MOTA$\uparrow$ & FP$\downarrow$ & FN$\downarrow$ & IDs$\downarrow$\\
		\midrule
		\midrule
		\multicolumn{7}{l}{\textit{MOT17 private detection}} \\
		DAN\cite{DAN} & 39.3 &49.5&  52.4& 25423& 234592& 8431 \\
		TubeTK\cite{tubetk} & 48.0 &58.6& 63.0 &27060 &177483& 4137 \\
		MOTR\cite{motr} & -& 66.4& 65.1 &45486& 149307 &2049 \\
		CTracker\cite{chained} & 49.0 &57.4 & 66.6 &22284 &160491 &5529 \\
		MAT\cite{Mat} & 53.8 &63.1 & 69.5& 30660 &138741& 2844 \\
		QuasiDense\cite{pang2021quasi} & 53.9 &66.3& 68.7 &26589 &146643& 3378 \\
		TransTrack\cite{sun2020transtrack} & 54.1 &63.5 &75.2 &50157 &86442 &3603 \\
		TransCenter\cite{xu2021transcenter} & 54.5 &62.2 & 73.2 &23112 &123738 &4614 \\
		GSDT\cite{wang2021joint} & 55.2& 66.5 &73.2 &26397 &120666 &3891 \\
		PermaTrackPr\cite{tokmakov2021learning} & 55.5 &68.9 & 73.8&28998 &115104 &3699 \\
		SOTMOT\cite{zheng2021improving} & - &71.9 & 71.0 &39537 &118983 &5184 \\
		FUFET\cite{shan2020tracklets} & 57.9 &68.0 &76.2 &32796 &98475 &3237 \\
		MTrack\cite{MTrack} & - & 72.1 & 73.5 &53361 & 101844 & 2028 \\
		FairMOT\cite{fairmot} & 59.3 &72.3 &73.7 &27507 &117477 &3303 \\
		CSTrack\cite{liang2022rethinking} & 59.3 &72.6 &74.9 &23847 &114303 &3567 \\
		SiamMOT\cite{siammot} & - &72.3 &76.3 &- &- &- \\
		ReMOT\cite{yang2021remot} & 59.7 &72.0 & 77.0&33204 &93612 &2853 \\
		Semi-TCL\cite{li2021semi} & 59.8 &73.2 &73.3 &22944 &124980 &2790 \\
		CorrTracker\cite{CorrTracker} & 60.7 &73.6 &76.5 &29808 &99510 &3369 \\
		RelationTrack\cite{relationtrack} & 61.0 &74.7 & 73.8&27999 &118623 &1374 \\
		TransMOT\cite{chu2021transmot} & 61.7 &75.1 &76.7 &36231 &93150 &2346 \\
		ByteTrack\cite{bytetrack}& 63.1 &77.3 & {\bf 80.3} &25491 &{\bf 83721} &2196 \\
		\rowcolor{red!15} \textbf{FineTrack} & {\bf 64.3} & {\bf 79.5} & 80.0 & {\bf 21750} & 90096 & {\bf 1272}\\
		\midrule
		\midrule
		\multicolumn{7}{l}{\textit{MOT20 private detection}} \\
		MLT\cite{zhang2020multiplex} & 43.2 &54.6& 48.9& 45660& 216803& 2187 \\
		TransTrack\cite{sun2020transtrack} & 48.5 &59.4 & 65.0&27197& 150197 &3608 \\
		FairMOT\cite{fairmot} & 54.6& 67.3& 61.8& 103440& 88901& 5243 \\
		Semi-TCL\cite{li2021semi} & 55.3 &70.1 & 65.2&61209 &114709& 4139 \\
		CSTrack\cite{liang2022rethinking} &54.0 & 68.6 & 66.6&25404& 144358 &3196 \\
		GSDT\cite{wang2021joint} & 53.6 &67.5 &67.1 &31913 &135409 &3131 \\
		SiamMOT\cite{siammot} & - &69.1 &67.1 &- &- &- \\
		RelationTrack\cite{relationtrack} & 56.5 &70.5 & 67.2 &61134 &104597 &4243 \\
		SOTMOT\cite{zheng2021improving} & - &71.4 &68.6 &57064 &101154 &4209 \\
		ByteTrack\cite{bytetrack} &61.3 &75.2 &77.8 &26249 &{\bf 87594} &1223 \\
		\rowcolor{red!15} \textbf{FineTrack} &{\bf 63.6}& {\bf 79.0} & {\bf 77.9}& {\bf 24439}& 89012& {\bf 980}\\
		\bottomrule
	\end{tabular}
	\caption{Comparison of the state-of-the-art methods under the private detection on the {\bf MOT17} and {\bf MOT20} test set. The best results are marked in {\bf bold} and our method is highlighted in \colorbox {red!15}{pink}.}
	\label{table:MOT17&MOT20}
\end{table}
\section{Experiments}
\subsection{Settings}
\noindent 
{\bf Datasets.} To verify the effectiveness of FineTrack, we test it on MOT17\cite{mot17}, MOT20\cite{mot20}, and DanceTrack\cite{dancetrack} datasets. MOT17 and MOT20 datasets provide training sets but do not contain validation sets. The test metric only can be obtained by uploading the tracking results to the MOTChallange website for evaluation. Therefore, in the ablation experiment phase, we divided the first half of each video of the MOT17 dataset into the training set and the second half as the validation set. Due to the difference between detection and Re-ID, we train them separately. For testing on MOT17, we train a detector using the same datasets as ByteTrack, including CrowdHuman\cite{crowdhuman}, MOT17, Cityperson\cite{Citypersons}, and ETHZ\cite{ETHZ}, and then freeze the trained detector parameters and train Re-ID separately on MOT17.
\par
DanceTrack is a large-scale multi-object tracking dataset for occlusions, frequent crossing, uniform appearance, and human body tracking with different body postures, with 100 videos. It uses 40, 25, and 35 videos as training, verification, and test set, respectively. The pedestrians in DanceTrack wear remarkably similar clothes, which is a massive challenge for Re-ID. Many existing methods that rely on Re-ID are at a disadvantage on this dataset.
\\
{\bf Metrics.} We use CLEAR-MOT Metrics\cite{metrics}, such as HOTA, MOTA, IDF1, IDs, FP, FN, etc., to evaluate the tracking performance. MOTA focuses on detection performance, and IDF1\cite{IDF1} emphasizes association performance. Compared with them, HOTA\cite{Hota} comprehensively balances detection, association, and localization effects. To accurately measure the performance of Re-ID and exclude the influence of detection and association, Rank-1 and mAP are utilized as metrics to measure the ability of Re-ID feature representation.
\\
{\bf Implementation details.} The partial feature and global feature dimensions are 128 and 256. The number of target masks is set as 6. The detector training Settings are consistent with Bytetrack. During training, the batch size is set to 8. The model parameters are updated using the Adam optimizer\cite{adam} with an initial learning rate of $2 \times {10^{ - 4}}$ and 20 training epochs. The learning rate is reduced to $2 \times {10^{ - 5}}$ at the $10_{th}$ epoch.

\begin{table} \small
	\centering
	\tabcolsep=1.4mm
	\begin{tabular}{@{}lccccc@{}}
		\toprule
		Method & HOTA$\uparrow$ & IDF1$\uparrow$ & MOTA$\uparrow$ & AssA $\uparrow$ & DetA $\uparrow$ \\
		\midrule
		CenterTrack\cite{centertrack}& 41.8 & 35.7& 86.8& 22.6 & {\bf 78.1}\\
		FairMOT\cite{fairmot} & 39.7 & 40.8& 82.2 &23.8 & 66.7\\
		QuasiDense\cite{pang2021quasi} & 45.7 & 44.8 & 83.0 & 29.2 & 72.1\\
		TransTrack\cite{sun2020transtrack} & 45.5& 45.2 & 88.4 &27.5 & 75.9\\
		TraDes\cite{wu2021track} & 43.3 & 41.2 & 86.2  &25.4  &74.5\\
		ByteTrack\cite{bytetrack} & 47.7 & 53.9 & 89.6  &32.1 & 71.0\\
		\rowcolor{red!15} \textbf{FineTrack} & {\bf 52.7} & {\bf 59.8} & {\bf 89.9} & {\bf 38.5} & 72.4\\
		\bottomrule
	\end{tabular}
	\caption{Comparison of the state-of-the-art methods on the {\bf DanceTrack} test set. The best results are marked in {\bf bold} and our method is highlighted in \colorbox {red!15}{pink}.}
	\label{table:DanceTrack}
	\vspace{-0.1cm}
\end{table}

\subsection{Comparison with the State-of-the-art Methods}
In this part, we compare the performance of FineTrack with previous SOTA methods on three benchmarks, i.e., MOT17, MOT20 and DanceTrack. Notably, some MOT methods utilize extra Re-ID models and Re-ID datasets to improve their performance of identity embedding. For a fair comparison, FineTrack only uses MOT datasets for Re-ID training and does not use any additional labels for supervision, such as masks, etc.
\\
{\bf MOT17:} FineTrack shows significant advantages without introducing extra Re-ID models and datasets containing a large amount of identity information. As shown in \cref{table:MOT17&MOT20}, FineTrack method achieves the best tracking accuracy on the MOT17 datasets (i.e. 64.3\% HOTA, 79.5\% IDF1 and 80.0\% MOTA, etc.). In cases where using detection simply can obtain excellent tracking accuracy, FineTrack achieves further improvement compared to BytaTrack (1.2\% HOTA and 2.2\% IDF1), which also adopts YOLOX as the detector. This demonstrates that our proposed fine-grained appearance representation dramatically improves the performance of identity embedding. Compared with other MOT methods, the advantages of FineTrack are more prominent. 
\\
{\bf MOT20:} Compared with MOT17, MOT20 is more crowded, which causes frequent and dense occlusion. As shown in \cref{table:MOT17&MOT20}, FineTrack outperforms ByteTrack by 2.3\% on HOTA, 3.8\% on IDF1 with superior identity embeddings while reducing IDs from 1223 to 980. FineTrack has a stable and excellent performance in dense pedestrian scenes, highlighting the role of fine-grained appearance embedding.
\\
{\bf DanceTrack:} Pedestrians in DanceTrack dress very similarly and have complex body pose variations, making appearance-based methods perform poorly. FineTrack has excellent appearance modeling capabilities, as shown in \cref{table:DanceTrack}, where we outperform ByteTrack entirely (5.0\% on HOTA and 5.6\% on IDF1, etc.). This further indicates the superiority of fine-grained appearance representations.

\begin{table}[t]
	\small
	\centering
	\tabcolsep=2.5mm
	\begin{tabular}{lcc}
		\toprule
		\multirow{2}*{Method} &
		\multicolumn{2}{c}{Metrics} \\
		\cmidrule{2-3}
		& Rank-1 & mAP  \\
		\hline
		1 FineTrack(train by the order of videos )    & 89.7 & 56.7\\
		2 FineTrack(shuffle all video frames)    & 90.9 & 60.4\\
		3 FineTrack(w SGS)  & {\bf 92.3} & {\bf 61.8}\\
		\toprule
	\end{tabular}
	\caption{{\bf Ablation study of SGS.} The first row does not disturb the video frames and trains in their order. The second row represents shuffling all the video frames. The last row is our SGS training strategy. (SGS: Shuffle-Group Sampling)} 
	\label{table:SGS}
\end{table}

\subsection{Ablation Studies}
In this section, we verify the effectiveness of FineTrack through ablation studies. All experiments are conducted on the MOT17 dataset. Since the MOT Challenge does not provide validation sets, we split the MOT17 dataset into two parts. The first half of each video serves as the training set and the second half as the validation set. To fairly measure the performance of identity embedding, experiments in \cref{table:SGS}, \cref{table:Components} and \cref{table:FAM} utilize the Ground Truth of bboxes as input, which can eliminate the interference caused by false detection. \cref{table:SGS} verifies the effectiveness of the SGS training strategy, which is employed in the remaining ablation studies.
\\
{\bf Analyze of SGS.} To validate the advantages of our proposed SGS training strategy, we compare it with two other typical training methods, as reported in \cref{table:SGS}. Our proposed SGS is superior to the first training method by 2.6\% on Rank-1 and 5.1\% on mAP. While shuffling all video frames gets improvement (1.2\% on Rank-1 and 3.7\% on mAP) over training in the order of video frames, there is still a gap of 1.4\% on both Rank-1 and mAP compared with SGS. This ablation experiment proves the effectiveness of FineTrack and reflects the importance of positive and negative samples balanced and discrete distribution of training data for training Re-ID.
\\
{\bf Component-wise Analysis.} In this part, we verified the effectiveness of FAFPN and MPMG through ablation experiments. As shown in \cref{table:Components}, FineTrack effectively improves the performance of appearance embeddings. Because YOLOX\cite{yolox} cannot extract Re-ID features, we use FPN\cite{FPN} to aggregate multi-scale feature maps and then adopt Global Average Pooling (GAP) to obtain appearance embeddings. Finally, we combine FPN and GAP as our baseline (row 1). Compared with baseline indicators, our method (row 4) outperforms the baseline by 4\% on Rank-1 and 11.4\% on mAP, demonstrating the value of exploring fine-grained appearance representation.
\par
\begin{table}[t]
	\small
	\centering
	\tabcolsep=1.4mm
	\begin{tabular}{lcccccc}
		\toprule
		\multirow{2}*{Method} &
		\multicolumn{4}{c}{Components} &
		\multicolumn{2}{c}{Metrics} \\
		\cmidrule(lr){2-5} \morecmidrules \cmidrule(lr){2-3} \cmidrule(lr){4-5} \cmidrule(lr){6-7}
		& FPN & FAFPN & GAP & MPMG & Rank-1 & mAP  \\
		\hline
		1 Baseline  & \checkmark &  & \checkmark &  & 88.3 & 50.4\\
		2  	    & \checkmark &  &  & \checkmark & 89.8 & 55.7\\
		3       &  & \checkmark & \checkmark &  & 91.8 & 58.2\\
		4 \textbf{FineTrack}  &  & \checkmark &  &\checkmark  & {\bf 92.3} & {\bf 61.8}\\
		\toprule
	\end{tabular}
	\caption{{\bf Ablation study of the components in FineTrack.} The first row is our baseline, and the last row is the metrics that can be obtained by adopting our proposed FAFPN and MPMG. (GAP: Global Average Pooling, FAFPN: Flow Alignment FPN, MPMG: Multi-head Part Mask Generator)} 
	\label{table:Components}
	\vspace{-0.1cm}
\end{table}

\begin{table}[t]
	\small
	\centering
	\tabcolsep=2.2mm
	\begin{tabular}{lcccc}
		\toprule
		\multirow{2}*{Method} &
		\multicolumn{2}{c}{Components} &
		\multicolumn{2}{c}{Metrics} \\
		\cmidrule(lr){2-3} \cmidrule(lr){4-5}
		& GAP & MPMG & Rank-1 & mAP  \\
		\hline
		1 FineTrack(w/o FAM)   & \checkmark &  & 91.7 & 57.9\\
		2 FineTrack(w/o FAM)   &  & \checkmark & 92.0 & 61.1\\
		3 FineTrack(w FAM)  & \checkmark &  & 91.8 & 58.2\\
		4 FineTrack(w FAM)  &  & \checkmark & {\bf 92.3} & {\bf 61.8}\\
		\toprule
	\end{tabular}
	\caption{{\bf Ablation study of FAM.} Re-ID embedding can be extracted through GAP or MPMG. We verify the effectiveness of FAM in these two cases. (FAM: Flow Alignment Module)} 
	\label{table:FAM}
	\vspace{-0.1cm}
\end{table}
Then, we analyze the effectiveness of each module separately. For multi-scale feature maps aggregation, replacing FPN with FAFPN (row 3) can achieve improvement (3.5\% on Rank-1 and 7.8\% on mAP) when generating Re-ID embeddings with GAP. Furthermore, when MPMG replaces GAP, using FAFPN (row 4) is better than using FPN (row 2) by 2.5\% on Rank-1 and 6.1\% on mAP. This indicates that FAFPN captures feature context information more effectively than FPN. At the same time, employing MPMG to extract Re-ID embeddings (row 2) is better than the baseline (1.5\% on Rank-1 and 5.3\% on mAP). When FAFPN replaces FPN, MPMG is 0.5\% and 3.6\% higher than GAP on Rank-1 and mAP. MPMG focuses on the discriminative details of the target and can describe the appearance of targets more comprehensively than GAP. 
\\
{\bf Analyze of FAM.} In this part, we conducted ablation experiments with or without FAM under the effect of GAP or MPMG, as reported in \cref{table:FAM}. Specifically, when GAP is used to generate Re-ID embeddings, employing FAM (row 1 and row 3) can increase 0.1\% on Rank-1 and 0.3\% on mAP. At the same time, if we adopt MPMG to extract Re-ID embeddings (row 2 and row 4), FAM can get the improvement (0.3\% on Rank-1 and 0.7\% on mAP), which is significantly better than adding FAM when using GAP. GAP is a global representation that aggregates all information indiscriminately. On the contrary, MPMG generates part masks to distinguish features explicitly and can guide FAM to perform pixel-level alignment more reasonably, even without mask label supervision.
\\
{\bf Analyze of our components when tracking.} We adopt the same association strategy to prove the effectiveness of different components in FineTrack when tracking. As shown in \cref{table:track}, using FAFPN or MPMG alone also improves the tracking performance. When combining these two modules, there is a significant advantage over the baseline (1.5\% on MOTA, 3.3\% on IDF1, and the IDs decreases from 253 to 115), demonstrating the effectiveness of FineTrack.

\begin{table}
	\centering
	\tabcolsep=2.5mm
	\begin{tabular}{@{}lccc@{}}
		\toprule
		Method & MOTA$\uparrow$ & IDF1$\uparrow$ & IDs$\downarrow$\\
		\midrule
		1 Baseline & 75.5 & 77.8 & 253 \\
		2 Baseline+FAFPN & 76.0 & 78.9 & 459 \\
		3 Baseline+MPMG & 76.4 & 79.7 & 451 \\
		4 Baseline+FAFPN+MPMG & {\bf 77.0} & {\bf 81.1} & {\bf 115} \\
		\bottomrule
	\end{tabular}
	\caption{{\bf Ablation study of our components when tracking.} The same association strategy is adopted for our proposed module to obtain the tracking results.}
	\label{table:track}
\end{table}

\section{Conclusion}
In this work, we have argued that diverse fine-grained representation is essential for MOT. However, existing methods focus on coarse-grained global features, which are extremely sensitive to noise. To effectively filter irrelevant information, we propose to explore diverse fine-grained appearance representation to obtain more comprehensive embeddings. As reported in the ablation studies, our presented FAFPN has great advantages in terms of aligning semantic and aggregating contextual information. Meanwhile, our constructed MPMG can effectively focus on different parts of the target without label supervision. In the training phase, we propose the SGS training strategy to improve the model performance effectively. We have verified the effectiveness of our proposed FineTrack on three public benchmarks (MOT17, MOT20, and DanceTrack) and achieved state-of-the-art performance. The experiment results indicate that diverse fine-grained representation can significantly improve the performance of Re-ID in MOT. We hope this work can be a new solution for appearance representation to generate discriminative identity embeddings.
\newpage
{\small
\bibliographystyle{ieee_fullname}
\bibliography{egbib}
}
\end{document}